%% file: main.tex
\documentclass[letterpaper, 10 pt, journal, twoside]{ieeetran}  
\IEEEoverridecommandlockouts                              
 
\usepackage[T1]{fontenc}
\usepackage{mathptmx} 
\usepackage{times} 
\usepackage{amsmath, amsfonts} 
\usepackage{mathtools}
\usepackage{amssymb}  
\usepackage{graphicx} 
\usepackage[ruled,vlined]{algorithm2e} 
\usepackage{booktabs}  
\usepackage{stfloats}
\usepackage{cite}
\usepackage{multirow}
\usepackage[hidelinks=false,bookmarks=false]{hyperref}  
\usepackage{multirow}
\usepackage{lineno}
\usepackage{colortbl}
\usepackage{tabularx} 
\usepackage{subfigure}
\usepackage{url}
\usepackage{bm}
\usepackage{caption}

\DeclareMathOperator*{\argmin}{\arg\!\min} 
\DeclareMathAlphabet{\mathcal}{OMS}{cmsy}{b}{n}
\DeclareMathAlphabet{\mathcal}{OMS}{cmsy}{m}{n}
\usepackage{balance}

\title{Enhancing Campus Mobility: Achievements and Challenges of Autonomous Shuttle ``Snow Lion''}

\author{Yingbing Chen, Jie Cheng, Sheng Wang, Hongji Liu, Xiaodong Mei, Xiaoyang Yan, Mingkai Tang, \\ Ge Sun, Ya Wen, Junwei Cai, Xupeng Xie, Lu Gan, Mandan Chao, Ren Xin, \\ Ming Liu$^{*}$, Jianhao Jiao$^{*}$, Kangcheng Liu, and Lujia Wang.
\thanks{$^{*}$ Ming Liu and Jianhao Jiao are corresponding authors.}
} 
\IEEEaftertitletext{
}

\begin{document} 

\maketitle

\input{secs/intro}
\input{secs/sys_framework_modules}
\input{secs/modules_perception}

\input{secs/modules_plan_control}
\input{secs/exp_evaluations}
\input{secs/lessons_conclusion}


\addtolength{\textheight}{0cm}   
 
\bibliographystyle{IEEEtran}
\bibliography{IEEEabrv,ref}

\vspace{0.5cm}  


\textbf{Yingbing Chen}, The Hong Kong University of Science and Technology, Clear Water Bay, Hong Kong, China. Email: ychengz@connect.ust.hk.

\textbf{Jie Cheng}, The Hong Kong University of Science and Technology, Clear Water Bay, Hong Kong, China. Email: jchengai@connect.ust.hk.

\textbf{Sheng Wang}, The Hong Kong University of Science and Technology, Clear Water Bay, Hong Kong, China. Email: swangei@connect.ust.hk.

\textbf{Hongji Liu}, The Hong Kong University of Science and Technology, Clear Water Bay, Hong Kong, China. Email: hliucq@connect.ust.hk.

\textbf{Xiaodong Mei}, The Hong Kong University of Science and Technology, Clear Water Bay, Hong Kong, China. Email: xmeiab@connect.ust.hk.

\textbf{Xiaoyang Yan}, The Hong Kong University of Science and Technology, Clear Water Bay, Hong Kong, China. Email: xyanaq@connect.ust.hk.

\textbf{Mingkai Tang}, The Hong Kong University of Science and Technology, Clear Water Bay, Hong Kong, China. Email: mtangag@connect.ust.hk.

\textbf{Ge Sun}, The Hong Kong University of Science and Technology, Clear Water Bay, Hong Kong, China. Email: gsunah@connect.ust.hk.

\textbf{Ya Wen}, The Hong Kong University of Science and Technology (Guangzhou), Guang Zhou, China. Email: yawen@hkust-gz.edu.cn.

\textbf{Junwei Cai}, The Hong Kong University of Science and Technology (Guangzhou), Guang Zhou, China. Email: junweicai@hkust-gz.edu.cn.

\textbf{Xupeng Xie}, The Hong Kong University of Science and Technology, Clear Water Bay, Hong Kong, China. Email: xxieak@connect.ust.hk.

\textbf{Lu Gan}, The Hong Kong University of Science and Technology, Clear Water Bay, Hong Kong, China. Email: lganaa@connect.ust.hk.

\textbf{Mandan Chao}, The Hong Kong University of Science and Technology (Guangzhou), Guang Zhou, China. Email: mchao549@connect.hkust-gz.edu.cn.

\textbf{Ren Xin}, The Hong Kong University of Science and Technology, Clear Water Bay, Hong Kong, China. Email: rxin@connect.ust.hk.

\textbf{Jianhao Jiao}, The Hong Kong University of Science and Technology, Clear Water Bay, Hong Kong, China. Email: jjiao@connect.ust.hk.

\textbf{Ming Liu}, The Hong Kong University of Science and Technology, Clear Water Bay, Hong Kong, China. Email: eelium@ust.hk.

\textbf{Kangcheng Liu} The Hong Kong University of Science and Technology (Guangzhou), Guang Zhou, China. Email: kangchengliu@hkust-gz.edu.cn.

\textbf{Lujia Wang} Shenzhen Institutes of Advanced Technology, Chinese Academy of Sciences, Shenzhen, China. Email: lj.wang1@siat.ac.cn.
  
\end{document}

%% file: secs/intro.tex
In recent years, the rapid evolution of autonomous vehicles (AVs) has reshaped global transportation systems. Leveraging the accomplishments of our earlier endeavor, particularly ``Hercules'' \cite{liu2021role}, an autonomous logistics vehicle for transporting goods, we introduce ``Snow Lion'', an autonomous shuttle vehicle meticulously designed to transform on-campus transportation, providing a safe and efficient mobility solution for students, faculty, and visitors. 

The main aim of this research is to improve campus mobility through a dependable, efficient, and eco-friendly autonomous transportation solution tailored to meet the diverse requirements of a university setting. This initiative significantly differs from the experiences of ``Hercules'' \cite{liu2021role}, as the campus environment presents a notable contrast to the structured environments of highways and urban streets. Emphasizing both security and passenger comfort, the primary focus is on passenger transportation. Achieving this goal involves a detailed examination of complex system designs that integrate trajectory planning adjustments, prioritizing pedestrian safety and acceptance. Moreover, the implementation encompasses supplementary devices and functionalities, including warning inattentive pedestrians or jaywalkers.

\begin{figure}[t] 
\centering
\includegraphics[width=1.0\linewidth]{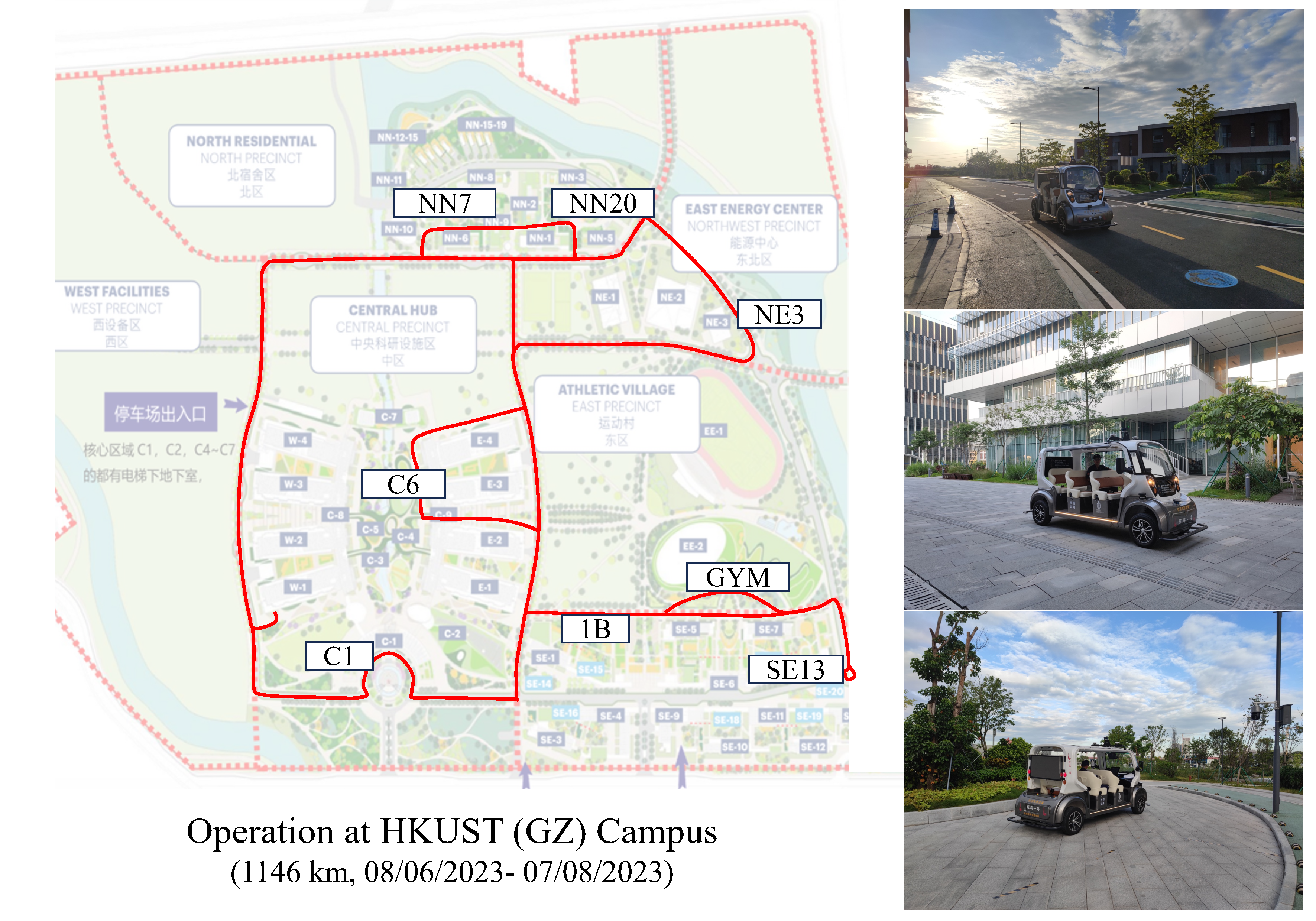} 
\caption{This figure illustrates the operational scenario of our autonomous shuttle during its service period at The Hong Kong University of Science and Technology (Guangzhou) (referred to as HKUST (GZ)). The red lines represent the operational road map of the campus, with shuttle stations depicted as boxes (e.g., C1 and 1B). The subfigures on the right exhibit multiple images captured during the operation of the autonomous shuttle.
}
\label{fig:front_page} 
\vspace{-1.5em}
\end{figure}

As a continuation of our previous work on ``Hercules'', this paper\footnote{The project website: \href{https://chenyingbing.github.io/xueshi_campus_av/}{\color{blue}{https://chenyingbing.github.io/xueshi\_campus\_av/}}.} contributes to the AVs for campus mobility. 
The development of this autonomous shuttle vehicle represents a significant stride toward a more intelligent, secure, and sustainable future for campus transportation. Furthermore, this paper rigorously explores the real-world challenges encountered during the developmental and implementation phases, as depicted in Fig. \ref{fig:front_page}. The experiments encompassed a $1146$-kilometer road haul and the transportation of $442$ passengers over a two-month period. It provides valuable insights into the intricate process of integrating an autonomous vehicle within campus shuttle operations. Additionally, a thorough analysis of the lessons derived from this experience furnishes a valuable real-world case study, along with recommendations for future research and development in the field of autonomous driving.


%% file: secs/sys_framework_modules.tex
\section{System Introduction}

\begin{figure*}[t!]
\centering
\includegraphics[width=\textwidth]{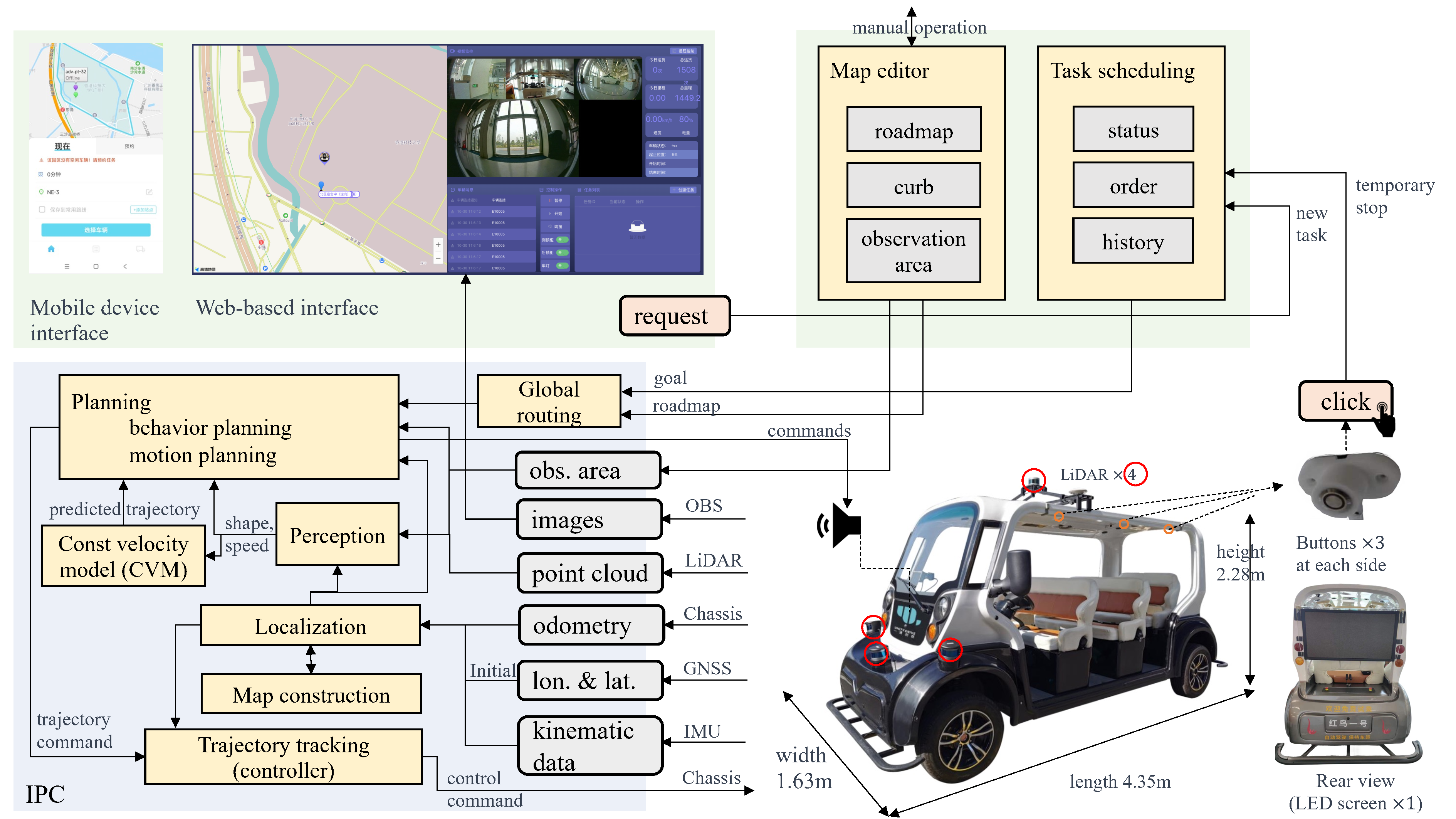}  
\caption{The functions and connections of the whole system of the autonomous vehicle. }
\label{fig_software_architecture}
\vspace{-1.0em}
\end{figure*} 

As shown in the Fig. \ref{fig_software_architecture}, we initially present an overview of the modules and their interconnections within our autonomous shuttle. Furthermore, we explore the contextual functions, onboard sensors, and elucidate their operational pipelines.

\subsection{Sensor and Device}

The autonomous shuttle, from a hardware perspective, boasts dimensions of $4350 \times 1630 \times 2280$ $\text{mm}^3$ and has the capacity to carry up to 6 passengers, with an maximum load capacity of 560 kilograms. The vehicle achieves a maximum driving speed of 15 km/h and is equipped with a navigation system that seamlessly integrates multi-sensor fusion for perception data, enabling circumstance map construction and localization functions. The platform houses an Industrial Personal Computer (IPC) equipped with an Intel i7-8700 CPU (6 cores and 12 threads) and 32 GB of memory, along with a 1050Ti NVIDIA graphics card. The vehicle is furnished with a removable 74V lithium-ion battery that provides power to the chassis, IPC, sensors, and accessories, enabling the vehicle to operate autonomously for 24 hours. 
Additionally, four 16-beam LiDAR units (HESAI XT16) are present: three are positioned at the front and sides to enhance the detection of surrounding obstacles, while one is installed on the roof primarily for localization purposes. 
The system also includes four fish-eye cameras utilized by OBS (Open Broadcaster Software) to record video data during operation, an IMU, and a high-precision GNSS (Global Navigation Satellite System) providing RTK (Real-Time Kinematic) capabilities, as well as latitude and longitude positioning information for the autonomous vehicle.
Finally, we have a 4G/5G Data Transfer Unit (DTU), six seat-side buttons, a sound player, and a rear LED screen at our disposal to bolster autonomous navigation. The DTU facilitates the connection of the Industrial Personal Computer (IPC) to our cloud management platform over the Internet, allowing for the initiation or suspension of navigation tasks via a mobile application. Additionally, seat-side buttons provide supplementary interaction opportunities for passengers on board during navigation. The sound player serves the dual purpose of broadcasting the vehicle's status and attracting the attention of nearby pedestrians who may be inattentive. Meanwhile, the LED screen is used for display purposes.

\subsection{Functional Pipeline}

The capabilities of the autonomous shuttle can be categorized into three primary parts: autonomous navigation, task scheduling, and various auxiliary modules, including map editing and sound broadcasting.

Tasks related to autonomous navigation primarily occur within the onboard IPC, involving perception, localization, and planning functions. The perception task interprets point cloud data from LiDARs, extracting geometric characteristics and velocity details of nearby entities. In contrast, the localization module employs this data for robot positioning and map creation. Once the shapes and speeds of other entities are acquired, their trajectories are estimated using a constant velocity model. These details, along with localization data, are then transmitted to the planning module to calculates a feasible trajectory for the AV along the given global route. Simultaneously, the planning module sends commands to the onboard sound system to enhance other road users' comprehension of the AV's movements. Ultimately, the controller tracks the obtained trajectory and generates a control command for the chassis to execute.

At a remote operational level, two primary tools: the mobile device interface and the web-based interface— are utilized for querying status, recording data, and issuing commands to the autonomous shuttle. Both interfaces offer comparable functions to users, harnessing identical data and services sourced from the cloud. The scheduling server handles task assignments and collects the status of all registered running vehicles. It also performs various functions such as accessing map data for routing, transmitting sensor data to the map server, recording key information in the log server, and providing data replay functionality for traceability.
From Fig. \ref{fig_software_architecture}, it is evident that for task scheduling, three methods are available for assigning tasks to unmanned vehicles. Firstly, developers can communicate with the scheduling server through web sockets. Secondly, during actual operations, users can utilize private smartphone applications to confirm their desired stops. Lastly, passengers can also indicate their intent to disembark by pressing an on-board button, prompting the scheduling module to arrange their exit at an appropriate location.


%% file: secs/modules_perception.tex
\section{Perception and Localization}

The LiDAR sensors collect point cloud data used in the perception and localization tasks of AVs. These functions are fundamental to the core of autonomous navigation. The perception function processes sensor data to comprehend the surrounding environment, while the localization function constructs environmental maps and provides location information.

\subsection{Multiple LiDAR-based 3-D Object Detection}

\begin{figure}[t]
\centering
\includegraphics[width=1.0\columnwidth]{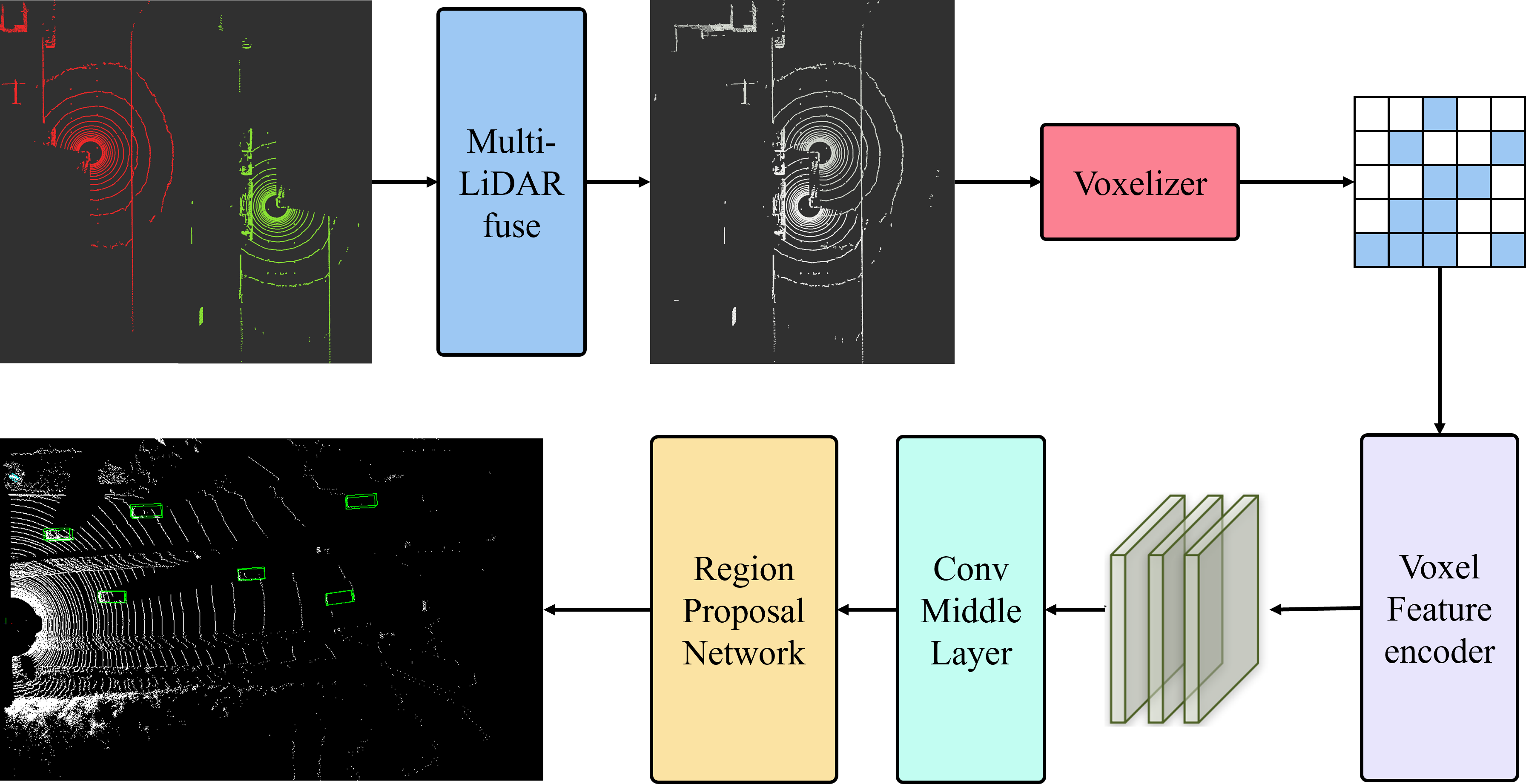}
\caption{The 3-D object detection module overview: Utilizing synchronized and well-calibrated LiDAR-captured point clouds, we employ an early-fusion technique to merge data from multiple calibrated LiDARs and apply VoxelNet \cite{zhou2018voxelnet} for 3-D object detection using the fused results.}
\label{fig_object_detection}
\vspace{-1.0em}
\end{figure}

The 3-D perception module empowers the autonomous vehicle to detect and precisely locate vital objects, including automobiles, pedestrians, and cyclists, in three-dimensional space via sensor data. 3-D object detection strives to both recognize and categorize objects while also estimating their positions and velocities relative to a designated coordinate system. Geometric data acquired by LiDAR sensors plays a crucial role in perception, as precise spatial information greatly enhances the accuracy of 3D object localization. 

\subsubsection{Multi-lidar Calibration} Multiple LiDAR sensors are used for object detection. The initial step involves calibrating the LiDAR sensors. In this study, we use our marker-based automatic calibration approach \cite{jiao2019novel} that eliminates the need for extra sensors and human involvement. In this approach, it is assumed that three linearly independent planar surfaces, arranged in a wall corner configuration, serve as calibration targets, ensuring that the geometric constraints are adequate for calibrating each pair of LiDAR sensors. Following the matching of corresponding planar surfaces, our approach effectively retrieves the unknown extrinsic parameters in two stages: initial estimation utilizing the closed-form Kabsch algorithm and subsequent refinement through plane-to-plane iterative closest point (ICP).

\subsubsection{Object Detection} 

An illustration of our 3-D object detection method is provided in Figure \ref{fig_object_detection}.	The VoxelNet approach \cite{zhou2018voxelnet} is employed to process multiple point clouds captured by multi-LiDAR sensors as inputs. During the input stage, an early-fusion scheme is utilized to combine data from multiple calibrated LiDAR sensors. Assuming that the LiDAR sensors are synchronized, we align all the raw point clouds to a common base frame before passing the fused point clouds to the 3-D object detector. Then, based on the given point cloud, we partition the 3D space into equidistantly spaced voxels. The point cloud information undergoes a systematic processing procedure, which commences with the mapping of raw point cloud data onto a three-dimensional voxel grid.	The voxelization step accomplishes two goals: it discretizes the continuous point cloud for computational manageability and addresses the issue of non-uniform point cloud distribution across voxels. For additional computational efficiency optimization, we employ a random subsampling strategy within each voxel, guaranteeing the capture of a representative data subset, thus mitigating the computational overhead linked to processing the complete voxel point cloud.	

The following stage, known as the Voxel Feature Encoding (VFE) layer, plays a crucial role in aggregating the key features of individual points within a voxel. The feature representation resulting from the VFE layer encapsulates the overall characteristics of the voxel. Additionally, the feature vectors in the vertical (z) direction are concatenated, effectively creating a bird's-eye view (BEV) of the current scene. This transformed representation is subsequently utilized for subsequent classification and regression tasks. 

At the feature point level, the region proposal network (RPN) header is crucial in predicting offset values for a specific set of anchors.	These offsets consist of eight values, including center coordinates, 3-D bounding box dimensions, orientation, and velocity. To enhance and optimize predictions while reducing redundancy, a non-maximum suppression (NMS) technique is employed to selectively retain predictions with the highest confidence scores. The RPN also plays a key role in classifying the object category associated with the current feature point, thereby enhancing the model's ability for comprehensive scene understanding. The resulting output comprises a series of 3-D bounding boxes, each bearing its associated category label, and includes their two-dimensional velocity in the horizontal plane.

\subsection{3-D Point-cloud Mapping and Localization} 

\subsubsection{Mapping}

\begin{figure}[t!] 
    \centering
    \begin{minipage}[t]{0.485\linewidth}
        \centering
        \includegraphics[width=\linewidth]{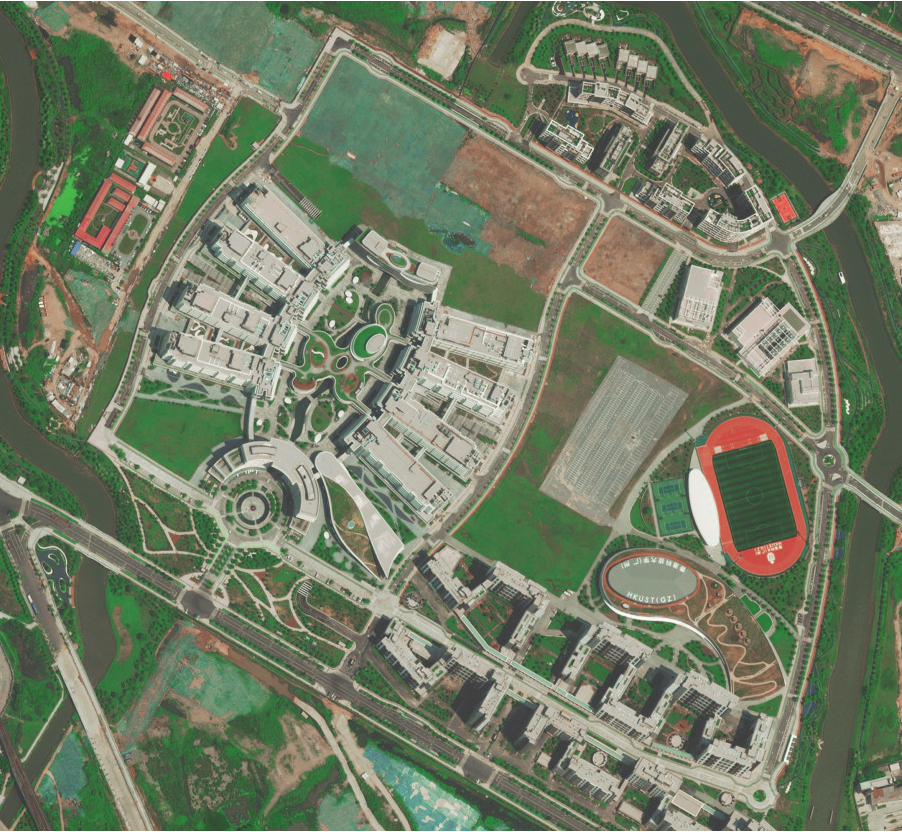}
        \label{fig: campus sm}
    \end{minipage}
    \begin{minipage}[t]{0.485\linewidth}
        \centering
        \includegraphics[width=\linewidth]{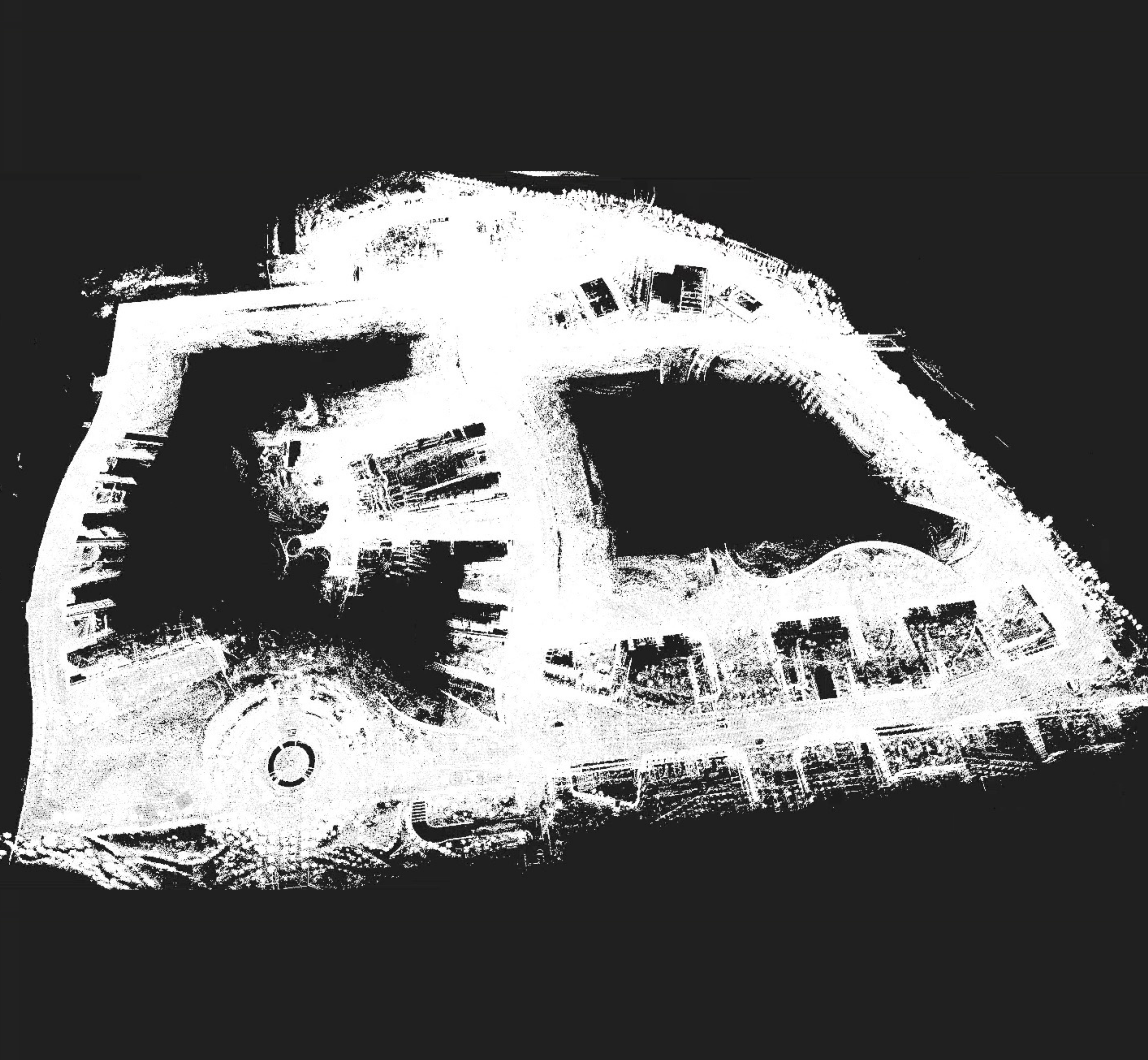}
        \label{fig: campus pc}
    \end{minipage}
    \caption{The figure displays a satellite map of the campus on the left and a constructed point cloud map on the right.}
    \label{fig: sm and bpt}
\vspace{-1.0em}
\end{figure}

In unmanned vehicle applications, maps play a pivotal role in providing essential information for the localization, perception, and planning tasks of unmanned vehicles. Among various map forms, LiDAR point cloud maps are preferred due to their density, informativeness, and accuracy. Graph-based optimization techniques are commonly employed for LiDAR-based mapping. Within a constructed pose graph, edges represent constraints, while nodes correspond to poses. The primary objective of pose graph optimization is to minimize the error associated with all constraints. LeGO-LOAM \cite{shan2018lego} is a widely recognized graph optimization method known for its lightweight design, ground optimization, backend enhancements, and loop detection mechanism. To further mitigate accumulation errors, we introduce GNSS measurements as a new constraint within the graph. The three constraints within the graph encompass the odometry constraint derived from LiDAR odometry, the GNSS constraint determined by the static transformation between the LiDAR and GNSS antenna, and the loop closure constraint established through the ICP algorithm. The optimization problem is addressed using the Levenberg-Marquardt (LM) algorithm, chosen for its effectiveness compared to the Gauss-Newton algorithm, owing to the LM algorithm's establishment of a trust region for valid non-linear approximations. Fig. \ref{fig: sm and bpt} depicts a map resulting from the mapping process alongside the corresponding real-world satellite map. The inclusion of GNSS measurements as a new constraint significantly enhances the accuracy and reliability of the optimized map. Our proposed approach holds promise for unmanned vehicle applications necessitating highly accurate and dependable maps.	

\label{sec: mapping}
\subsubsection{Localization}

Accurate localization is a critical aspect of navigation and control. Our approach to achieve precision localization involves fusing LiDAR observation data with odometry data. To relieve the issue of observation data delays, our localization system is designed based on the steady-state approximation of the Extended Kalman Filter (SSKF \cite{valls2018design}), as opposed to using a pure EKF. Upon acquiring a new frame of LiDAR data, it is imperative to register it with existing map data. In an effort to enhance computational efficiency and reduce storage demands, our initial step involves downsampling the map created through the method described in Sect. \ref{sec: mapping}. After acquiring an initial pose estimation using GNSS data, we proceed to extract the point cloud within the region of interest (ROI) from the map. Subsequently, we employ the ICP algorithm to calculate the precise pose.

%% file: secs/modules_plan_control.tex
\section{Planning Framework}

\begin{figure*}[!t]
\centering
\includegraphics[width=1.0\linewidth]{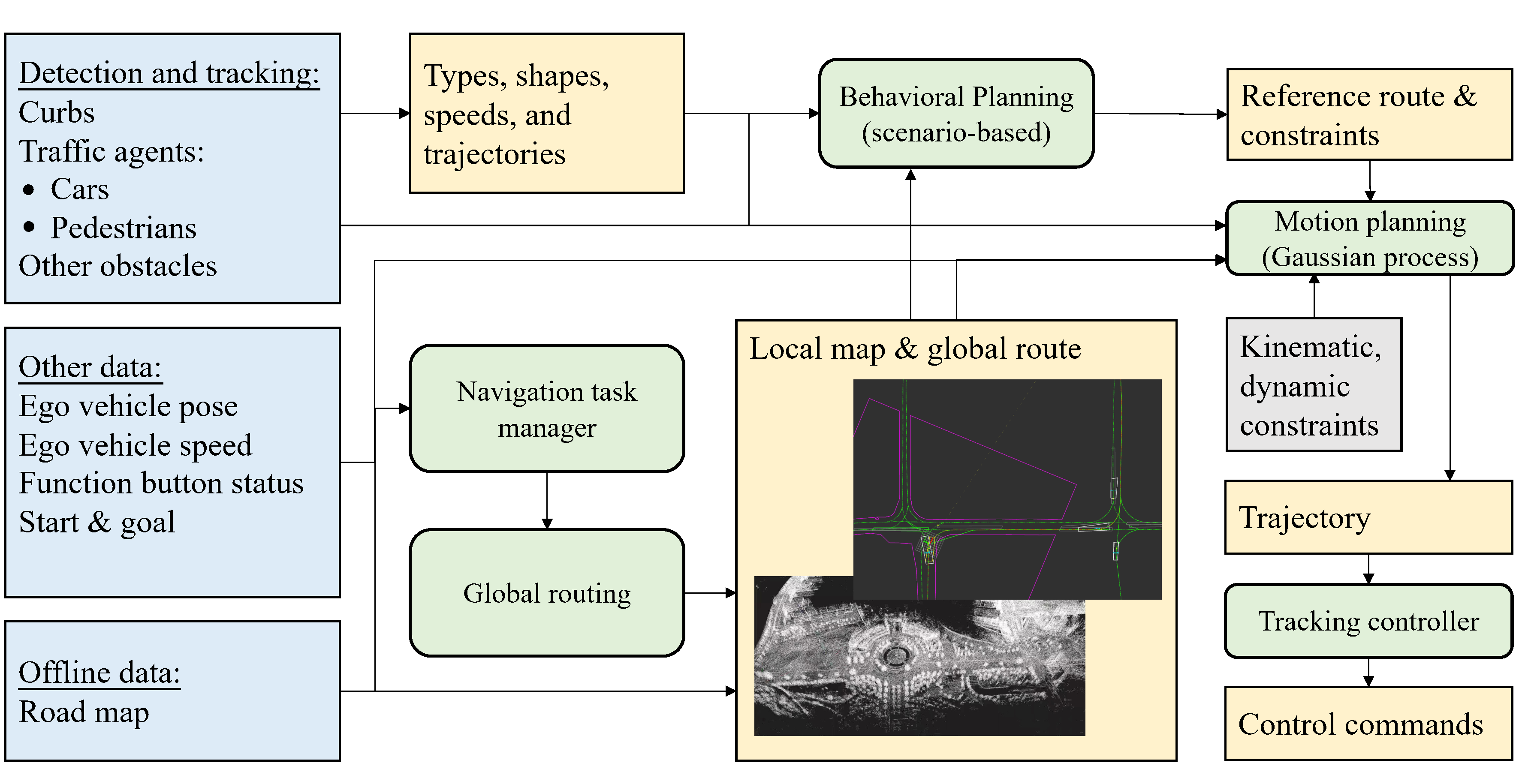}
\caption{The planning pipeline of the AV, mainly consisting of four parts: global routing, behavioral planning, motion planning, and tracking controller. }
\label{fig:planning_framework}
\vspace{-0.0em}
\end{figure*} 

In a complex campus environment, motion planning for an autonomous campus mobility vehicle must address diverse interactions, involving pedestrians, bicycles, and vehicles, both within and outside traffic-regulated zones. These demands necessitate the AV to possess two crucial capabilities. Firstly, the AV must demonstrate robust responses to environmental uncertainties. Second, it must ensure motion that is easily understood by other road users and comfortable for passengers. Fig. \ref{fig:planning_framework} illustrates our approach, which utilizes a pipeline planner for campus mobility scenarios.

\subsection{Navigation Task with Global Routing}
Building upon our prior research \cite{liu2021role, chen2022efficient}, when provided with the road map data and the AV's initial position, the navigation task manager manages destination data for the AV and invokes the global routing module to determine the global route for autonomous driving. As depicted in Fig. \ref{subfig:routing_localmap}, we apply A* algorithm to perform this function. Additionally, our designed task manager utilizes the status of function button to truncate the acquired global route to a temporary destination when passengers request an unscheduled drop-off. After the AV has come to a halt, it remains stationary for a predefined period (typically 8 seconds) and checks for the absence of nearby pedestrians before resuming the initial navigation task. 

\subsection{Behavioral Planning \label{subsect:behavioral_planning}}

\begin{figure}[t]
\centering
\includegraphics[width=1.0\columnwidth]{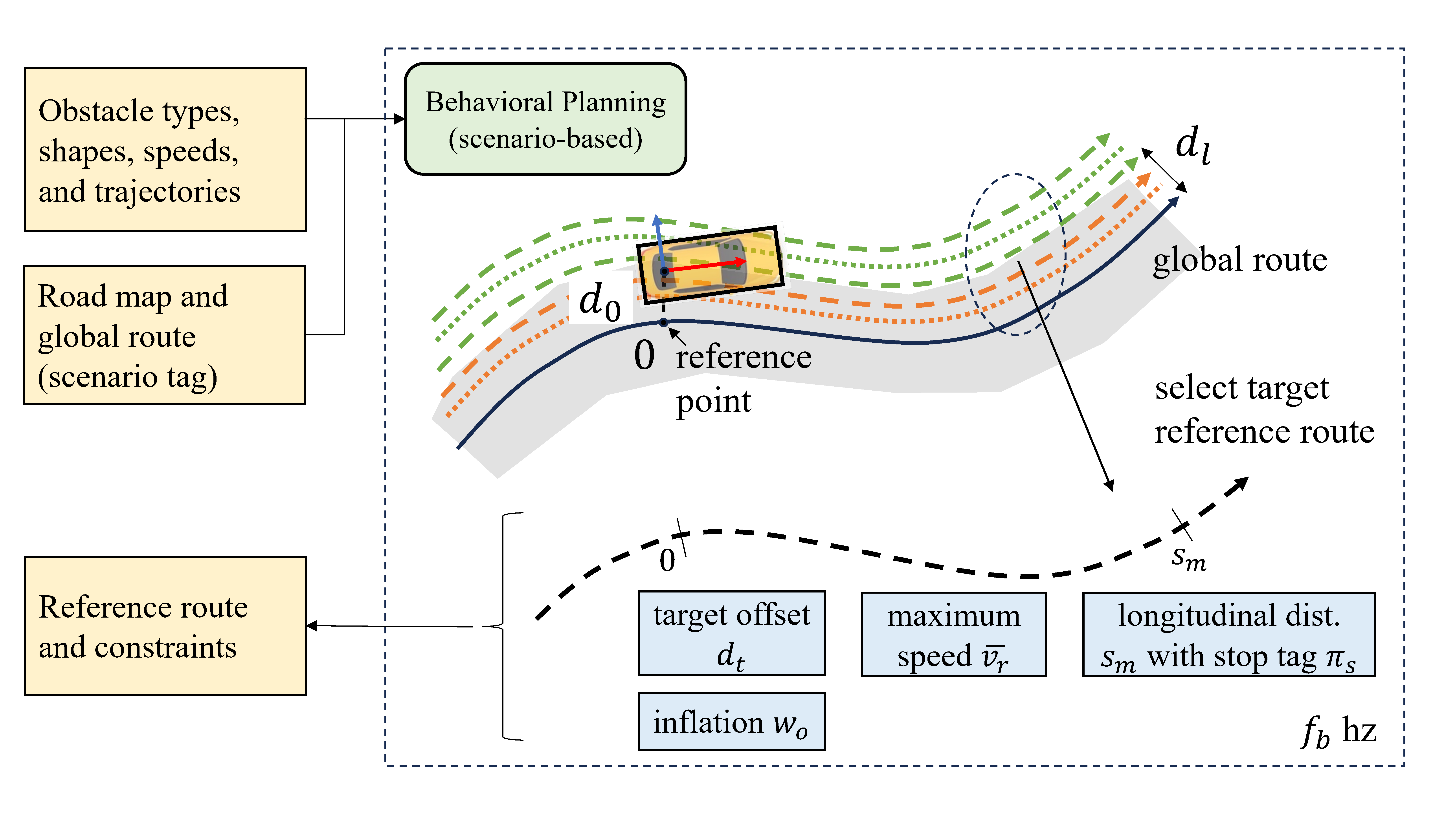}
\caption{An illustration of the behavioral planning, where $d_0$ represents the initial lateral offset of the AV (yellow block).}
\label{fig:behavior_plan}
\vspace{-1.0em}
\end{figure} 

Utilizing input data from the local map (representing the drivable area) and obstacle information, our behavioral planning, depicted in Fig. \ref{fig:behavior_plan}, computes a local reference route for subsequent motion planning tasks based on scenario tags in the road map. The resulting reference route is depicted as a lateral deviation $d_r$ from the global reference route, using the Fren\'et Framework. 
The goal of this module is to ensure a comfortable driving experience and define precise driving constraints for diverse scenarios. These scenarios encompass common roads, parking lots, and unprotected intersections. We specify distinct configurations and constraints for each scenario, subsequently integrating them into the motion planning task to ensure safety and stability during vehicle operation.

\subsubsection{Common situation} In the majority of cases, behavioral planning facilitates lane-changing maneuvers for autonomous vehicles, allowing the sampling of lateral offsets exceeding a specified threshold value $d_l$. Illustrated in Fig. \ref{fig:behavior_plan}, in such instances, the lateral offsets of the reference route are constrained within the green parts of lateral offsets, and a time delay $1/f_p$ is implemented in the planning loop to mitigate the occurrence of serpentine vehicle movements, thus ensuring a more comfortable passenger experience.

\subsubsection{Parking lot} Within parking lot scenarios, the environment becomes notably more uncertain due to the possibility of a parked car unexpectedly entering the AV's path. To ensure safe navigation performance, we employ both the predicted trajectories of other traffic agents and a width inflation operation to enhance driving safety.

\subsubsection{Unprotected intersections} During traversal of unprotected intersections, the behavioral planner prioritizes the consideration of dynamic obstacles within the corresponding observation areas (as depicted in Fig. \ref{subfig:routing_localmap}). When the autonomous vehicle detects dynamic obstacles within these zones, it performs a lane-stop maneuver, halting at a pre-defined stop line if their velocities and anticipated paths indicate potential future conflicts. This operation ensures stable driving performance, especially in intricately unprotected intersections, as relying solely on noise-affected trajectory predictions for planning purposes are generally inadequate.


This study utilizes the scenario tag of the global route to match predefined lateral offset options during the behavioral planning stage. Additionally, it determines parameters such as the obstacle width inflation volume $w_o$ and the maximum reference speed $\overline{v}_r$. Details of the configuration and constraint outputs will be elaborated in Sect. \ref{sect:experiments}. The target reference route is derived by computing the cost trajectory among the offset candidates $\mathcal{D}$, as
\begin{equation*}
d_t = \argmin\limits_{d \in \mathcal{D}}{(J_s + J_d + J_o + J_{\text{dyn}})},
\end{equation*}
\noindent where $J_s = w_s \cdot \left(1.0 - {s_m(d)}/{s_{\text{max}}}\right)$ promotes the achievement of an extended uncollided path distance. Here, $s_m(d)$ represents the route distance at offset $d$, while $s_{\text{max}}$ denotes the maximum sample distance. $J_d = w_{d1} |d - d_0| + w_{d2} |d|$ encourages the alignment of the target offset $d_t$ with both the route center and the AV's current offset value $d_o$. The expression for $J_o$ is given as ${w_{o1}}/{c_{\text{avg}}} + {w_{o2}}/{c_{\text{min}}}$, where $w_{(\cdot)}$ represents the weights, $c_{\text{avg}}$ denotes the average clearance to obstacles along the reference route,  $c_{\text{min}}$ represents the minimum clearance value. Additionally, $J_{\text{dyn}}$ is a cost term that assesses potential collisions with dynamic obstacles, incurring a high cost in the presence of collision risks with other agents' trajectories. Furthermore, the ultimate output reference route is labeled with $\pi_s$, signifying whether it undergoes truncation as a result of colliding with a static obstacle, at which point the AV must come to a stop at the end of the reference route.


\subsection{Motion Planning and Control} 

Upon acquiring the reference route and constraints from the behavioral planning module, the subsequent objective is to create the trajectory for the autonomous vehicle following the specified reference route.

\subsubsection{Constraints}

Apart from the maximum speed limit $\overline{v}_r$ of the reference route, the motion planning module encompasses various constraints: i) the speed limit influenced by the path curvature; ii) the speed restriction related to pedestrian clearance, considering the obstacle's width inflation $w_o$; and iii) the terminal speed limit if a halt is necessary at the end of the reference route. For the speed limit from path curvature, it follows 
\begin{equation}
    \overline{v}_{\kappa}(s) \leq \sqrt{\overline{a}_{\mathrm{lat}} / |\kappa(s)| },
\label{eq:vlimit_curvature}
\end{equation}
\noindent where $\overline{a}_{\mathrm{lat}}(s)$  represents the maximum lateral acceleration limit at the position $s$, while $\kappa$ symbolizes the curvature. Speed constraints due to pedestrian clearance are depicted through piecewise curves, following
\begin{equation}
\overline{v}(c) = 
\begin{cases}
v_{\text{min}} &\text{if } c < \delta_{\text{min}}, \\
\text{lerp}(v_{\text{min}}, v_{\text{mdn}}, c) &\text{elif } c < \delta_{\text{mdn}}, \\
\text{lerp}(v_{\text{mdn}}, v_{\text{max}}, c) &\text{elif } c < \delta_{\text{max}}, \\
v_{\text{max}} &\text{else }, \\
\end{cases}
\label{eq:vlimit_pedestrian}
\end{equation}
\noindent where $\text{lerp}(\cdot)$ represents the linear interpolation function. The variable $c$ represents the clearance to pedestrians at a specific sampling point, taking into account the shape of the AV. The symbol $\delta_{(\cdot)}$ signifies predefined clearance thresholds, while $\overline{v}{(\cdot)}$ represents the associated speed limits.

\subsubsection{Motion planning}

The objective of motion planning is to generate a trajectory for the autonomous vehicle that is both safe and kinodynamically feasible, in accordance with the decision made by the behavior planner. We utilize an iterative, path-speed decoupled method. First, a local Frenét coordinate frame is constructed based on the reference route, and the surrounding static obstacles are projected onto this frame. Subsequently, we structure the path generation problem using our Gaussian Process Motion Planning (GPMP) framework \cite{cheng2022real}. This framework converts collision and speed limit constraints (refer to (\ref{eq:vlimit_curvature}) and (\ref{eq:vlimit_pedestrian})) into probabilistic factors. The resultant path is derived from a maximum a posteriori problem. Upon establishing the initial path, predicted trajectories of dynamic obstacles are projected relative to this path, and an s-t graph is formulated. We then utilize a breadth-first search on this graph to identify an initial speed profile, which is later refined using piecewise polynomials through a quadratic programming problem. By integrating the path and speed profile, we assess the kinodynamic feasibility of the trajectory. If the resultant trajectory is found to breach any constraints, additional curvature constraints—sourced from the current speed profile—are incorporated into the path generation procedure, prompting the generation of a new path. Likewise, a fresh speed profile is constructed based on this new path. This iterative process continues until the trajectory either satisfies all constraints or exceeds the stipulated time limit.  

\subsubsection{Control}
The controller plays a crucial role in ensuring the safe operation of the vehicle, especially when dealing with varying road conditions. Therefore, we develop a controller to track the trajectory generated by the upstream motion planner accurately and robustly. Since our vehicle is based on Ackermann steering, the vehicle model can be simplified to a simpler bicycle model.
Traditional controllers often decouple the lateral and longitudinal motion of the vehicle model to simplify the controller design. While this simplification may ease the design process, it often sacrifices control accuracy. In contrast, our controller takes into account the coupled lateral and longitudinal dynamics of the vehicle model, which allows us to simultaneously control the steering angle and acceleration of the vehicle, resulting in improved tracking accuracy.
To ensure the robustness and safety of the vehicle, we design our controller based on Model Predictive Control (MPC) \cite{rajamani2011vehicle}, which models vehicle kinematics and dynamics, and then computes optimal control outputs according to predefined constraints and loss function. The loss function incorporates various factors such as lateral error, longitudinal error, heading angle error, and other tracking errors derived from the planner's given state. By carefully tuning the parameters of the controller, our vehicle can execute accurate trajectory tracking under different road conditions on the campus.

%% file: secs/exp_evaluations.tex
\section{Experimental Results \label{sect:experiments}} 

\begin{table}[t]
\renewcommand\arraystretch{1.15}   
\renewcommand\tabcolsep{8pt}  
\centering
\caption{The schedule specifies the autonomous shuttle's two-hour morning departure, with station locations provided in Fig. \ref{fig:front_page}. Similar operational shifts take place at noon (12:00 - 14:00) and in the evening (17:30 - 19:00).}
\begin{tabular}{p{4.5mm}|p{4.5mm}|p{4.5mm}|p{4.5mm}|p{4.5mm}|p{4.5mm}|p{4.5mm}|p{4.5mm}}
    \hline
    NN7 & NN20 & NE3 & C6 & SE13 & GYM & 1B & C1  \\ \hline
    08:00 & 08:03 &  & 08:10 &  &  &  &   \\ \hline
    08:20 & 08:23 &  & 08:30 &  &  &  &   \\ \hline
    08:40 & 08:43 &  & 08:50 & 09:02 & 09:05 & 09:06 & 09:13 \\ \hline
     &  &  &  &  09:25 & 09:28 & 09:29 & 09:36   \\ \hline 
     &  &  &  &  09:48 & 09:51 & 09:52 & 10:00   \\ \hline
\end{tabular}
\label{table:shuttle_schedule}
\vspace{-0.0em}
\end{table}

\begin{figure*}[t]
    \begin{minipage}[b]{1\linewidth}
        \subfigure[]{\includegraphics[width=0.66\linewidth]{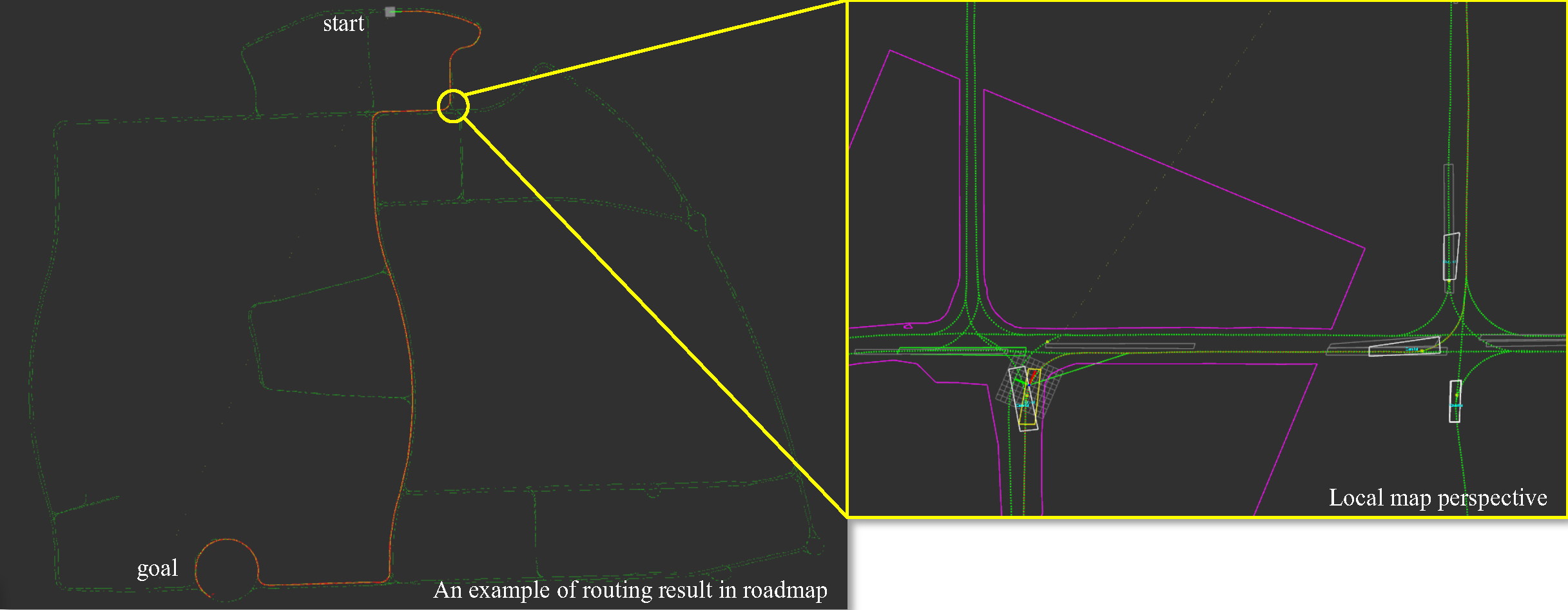}\label{subfig:routing_localmap}}\hspace{4.5pt}\subfigure[]{\includegraphics[width=0.33\linewidth]{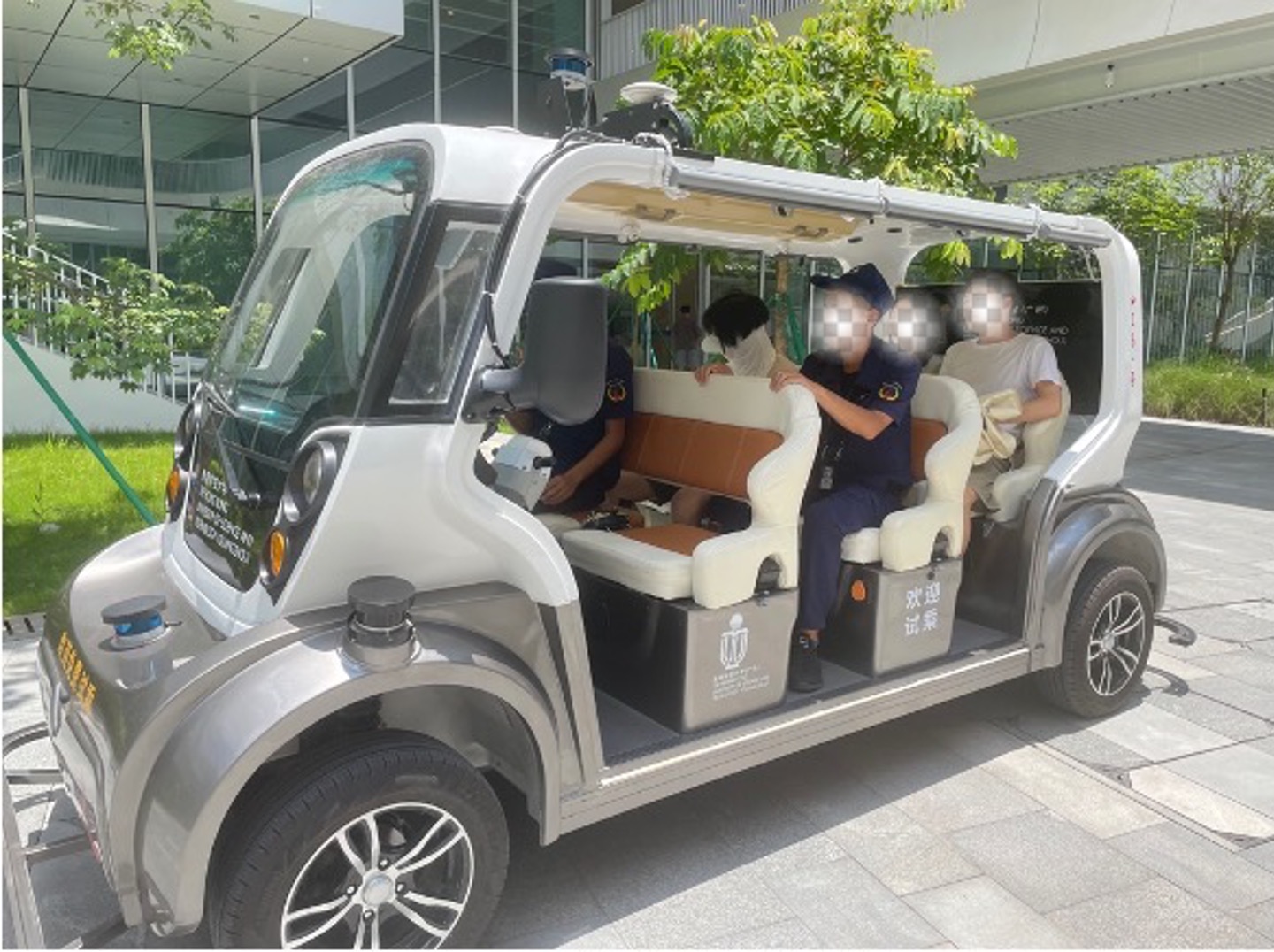}}
    \end{minipage} \vspace{-2mm} 
    \\
    \begin{minipage}[b]{1\linewidth}
	\subfigure[]
        {\includegraphics[width=0.33\linewidth]{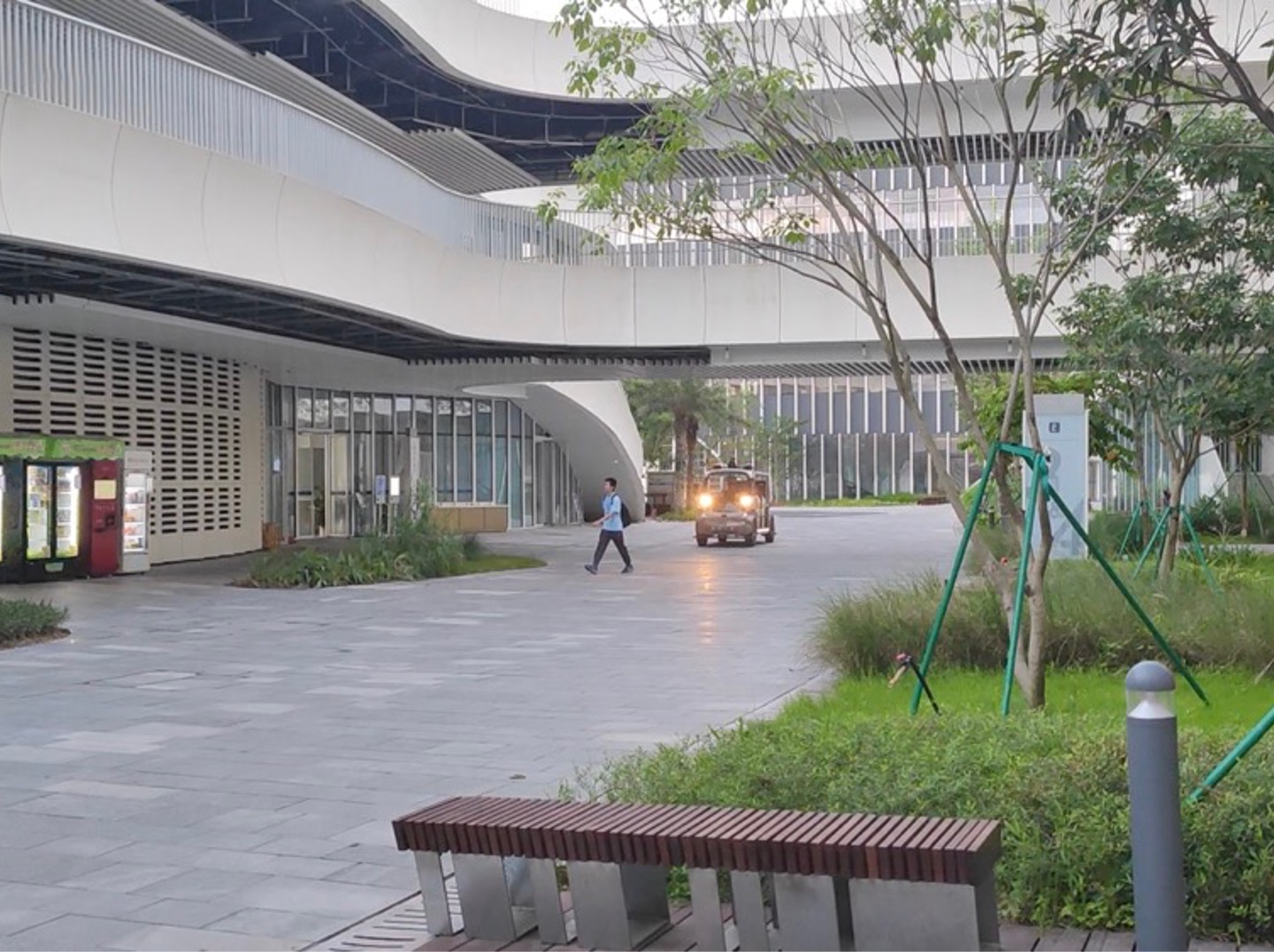}}\hspace{2.5pt}\subfigure[]{\includegraphics[width=0.33\linewidth]{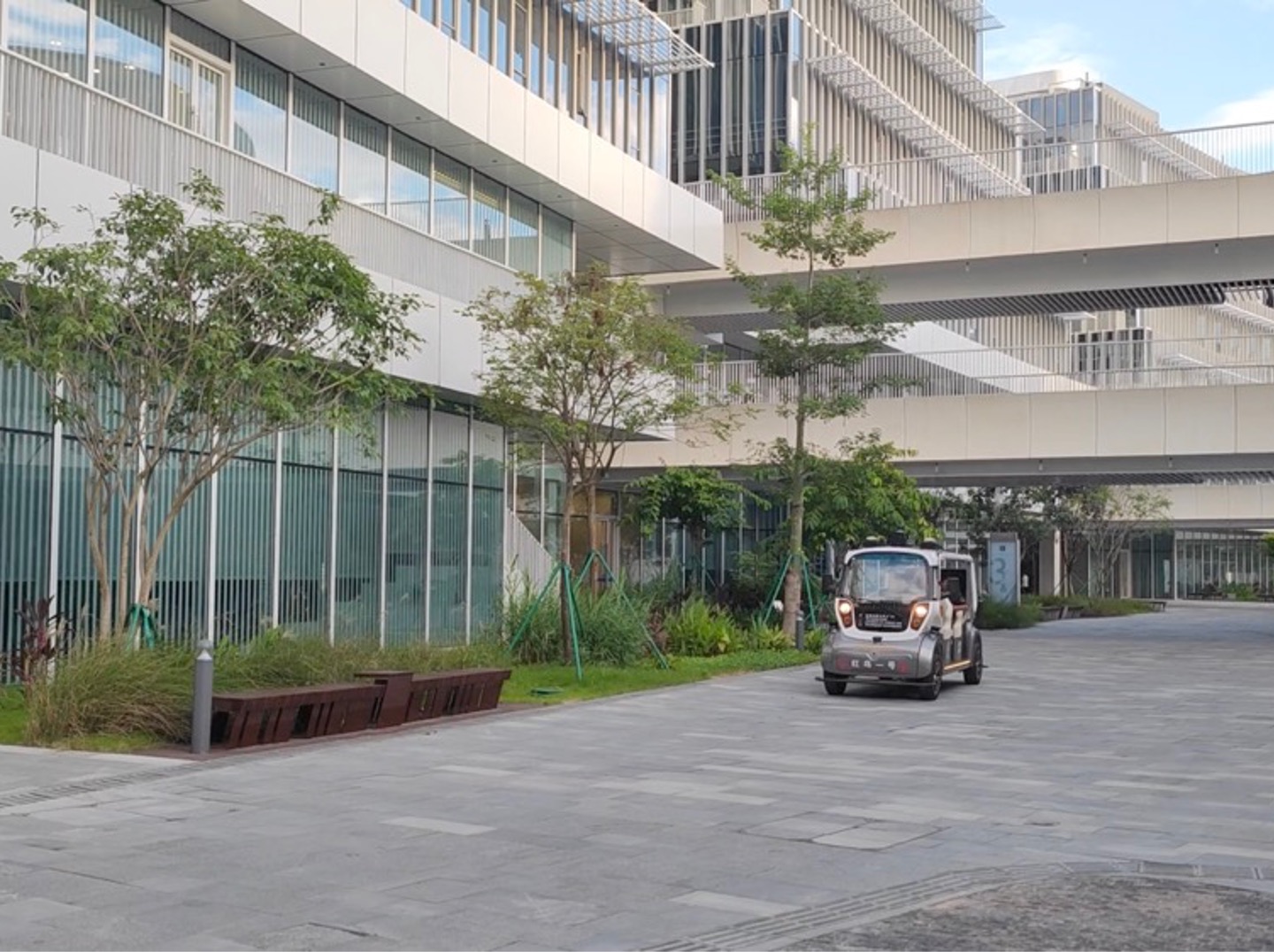}}\hspace{2.5pt}\subfigure[]{\includegraphics[width=0.33\linewidth]{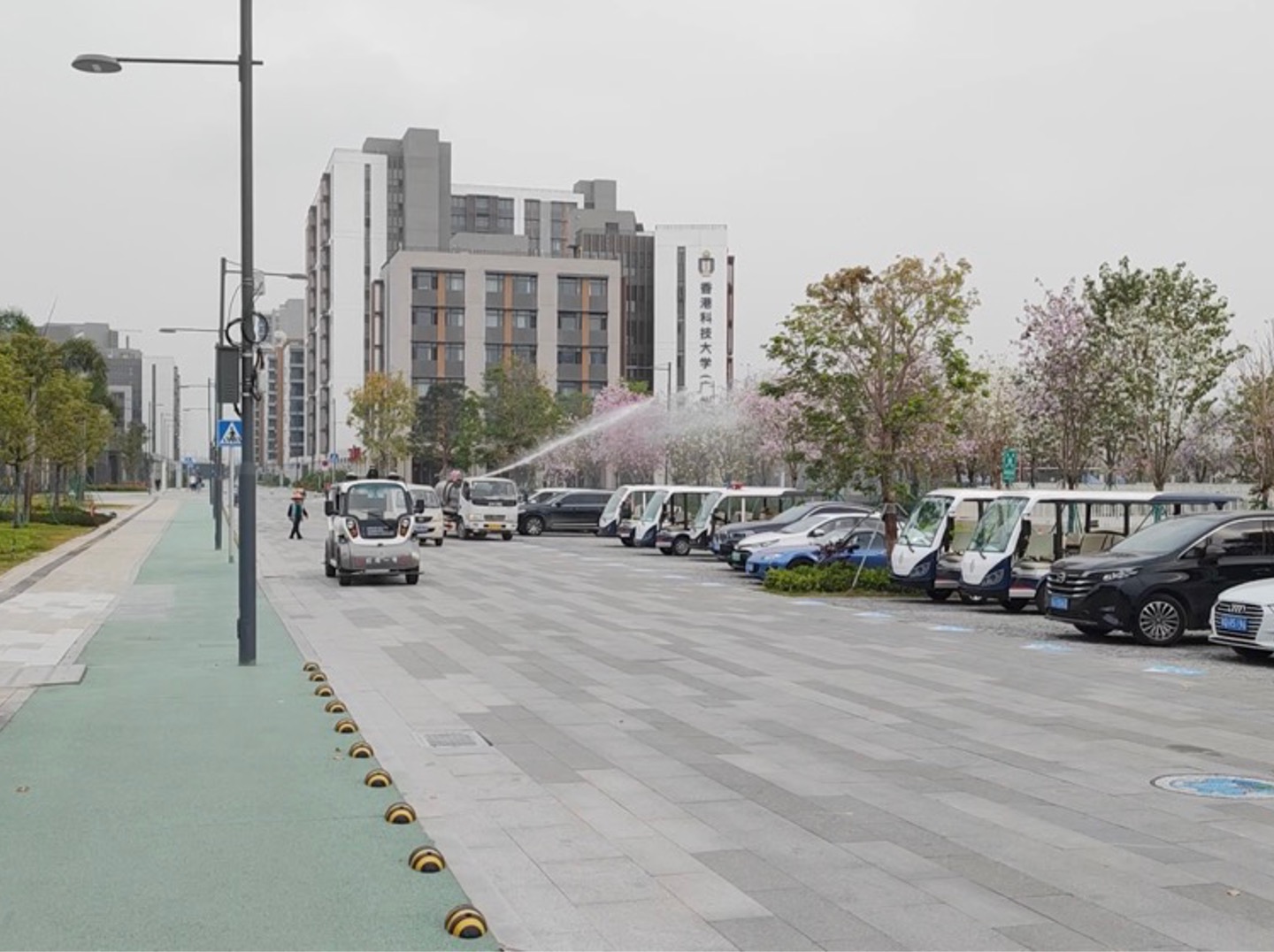}}
    \end{minipage} \vspace{-2mm} 
    \\ 
    \begin{minipage}[b]{1\linewidth}
	\subfigure[]
        {\includegraphics[width=0.33\linewidth]{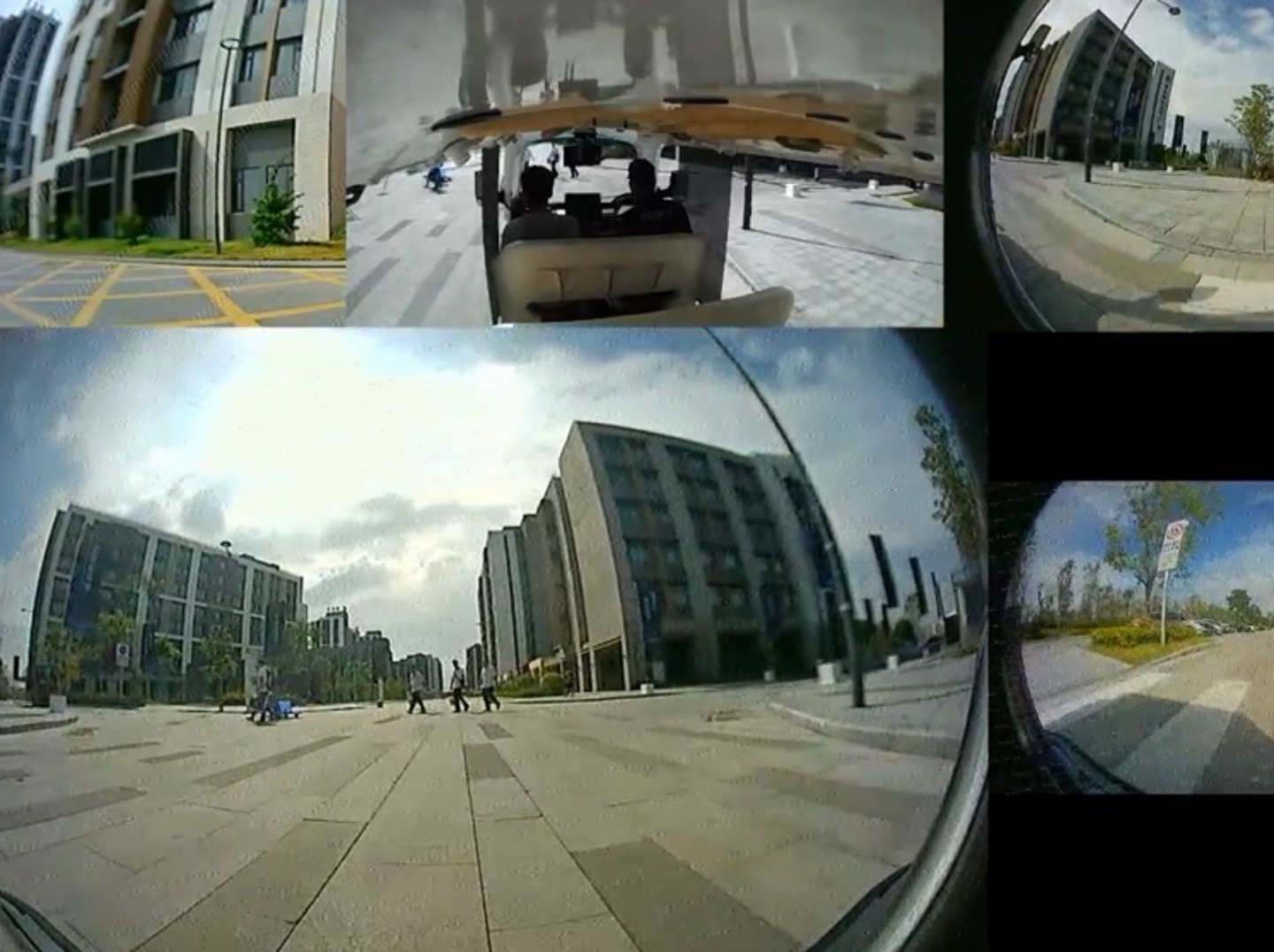}}\hspace{2.5pt}\subfigure[]{\includegraphics[width=0.33\linewidth]{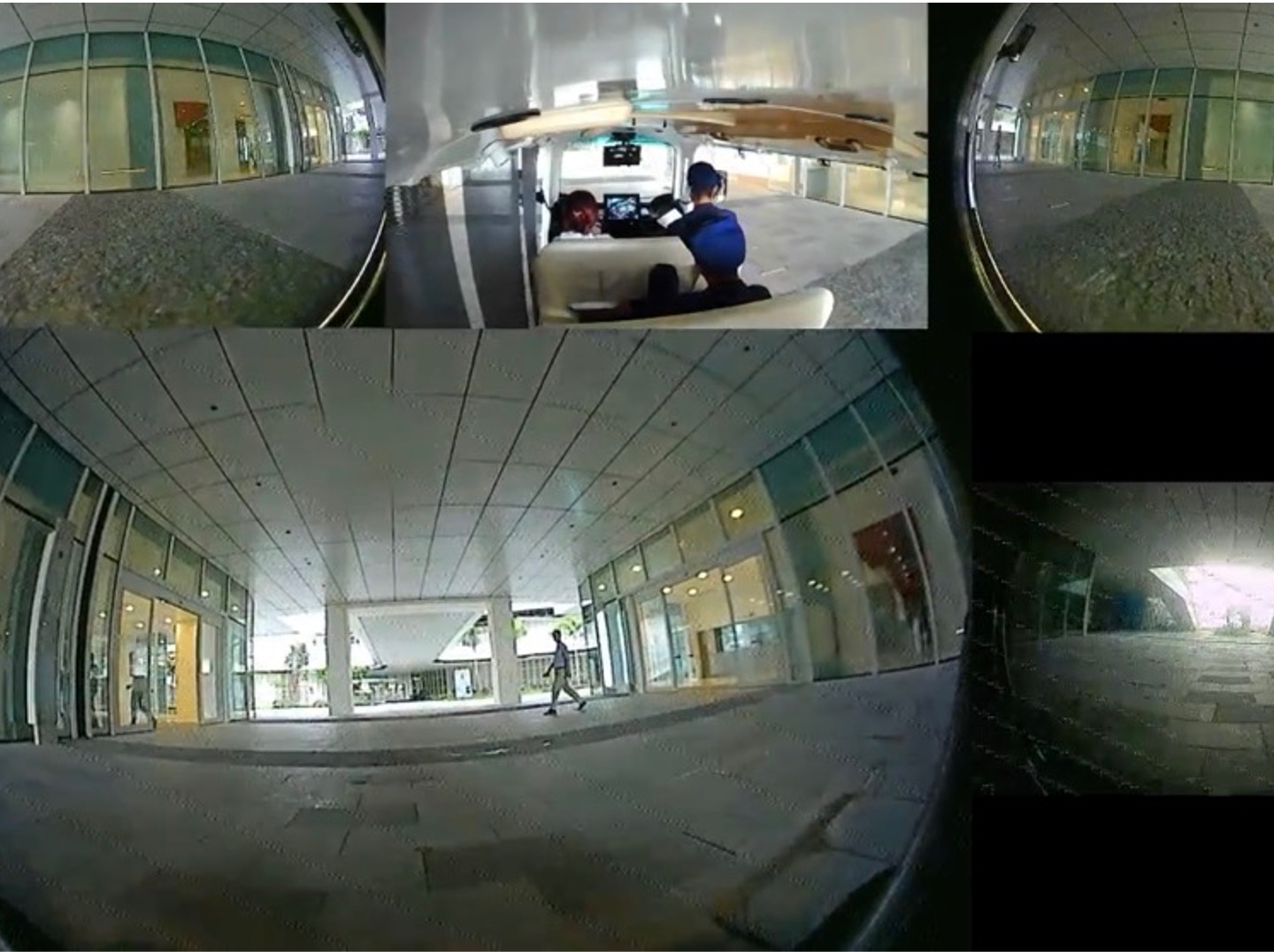}}\hspace{2.5pt}\subfigure[]{\includegraphics[width=0.33\linewidth]{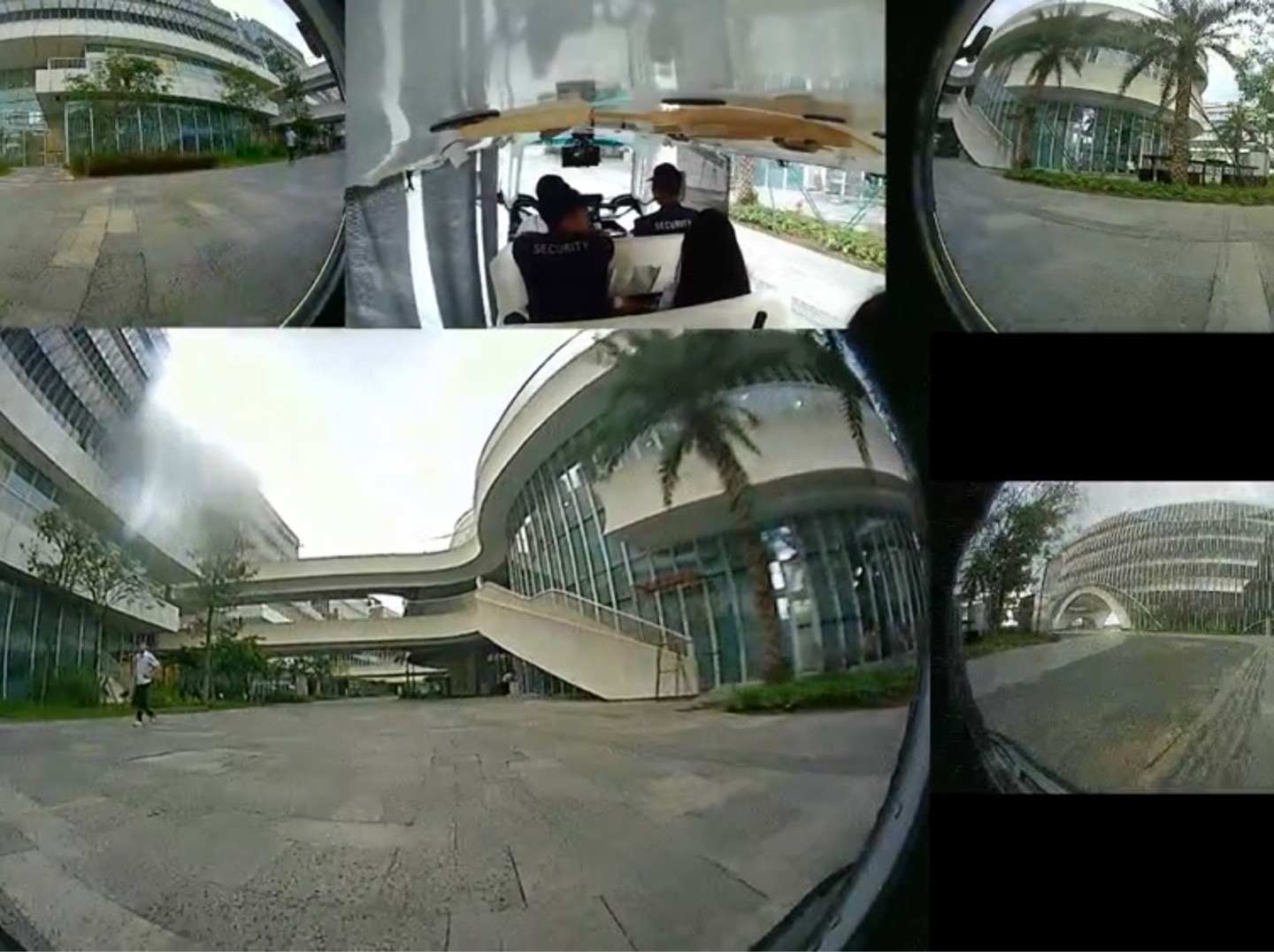}}
    \end{minipage} \vspace{-2mm}
    \caption{This figure presents examples of routing outcomes and operational photographs: (a) Examples of routing results and the local map perspective. The red line on the left represents the global route, while the right side displays local environmental representations used in the planning process. Purple boxes signify predefined curb areas. Green boxes designate the observation area utilized in behavioral planning (see Sect. \ref{subsect:behavioral_planning}). The yellow block indicates the stop line where the AV yields to dynamic obstacles within the intersection. (b)  The autonomous shuttle is stationed at the designated stop, awaiting passengers (two are already seated in the rear row). Two security guards dressed in blue are present throughout the entire operation to ensure passenger safety in case of accidents or emergencies. (c-e) Photographs captured during the operation period. (f-h) Screenshots obtained from the OBS during the operation period.}
    \label{fig:real_demos}
\end{figure*}
\begin{figure*}[t]
    \begin{minipage}[b]{1.0\linewidth}
        \subfigure[]{\includegraphics[width=0.66\linewidth]{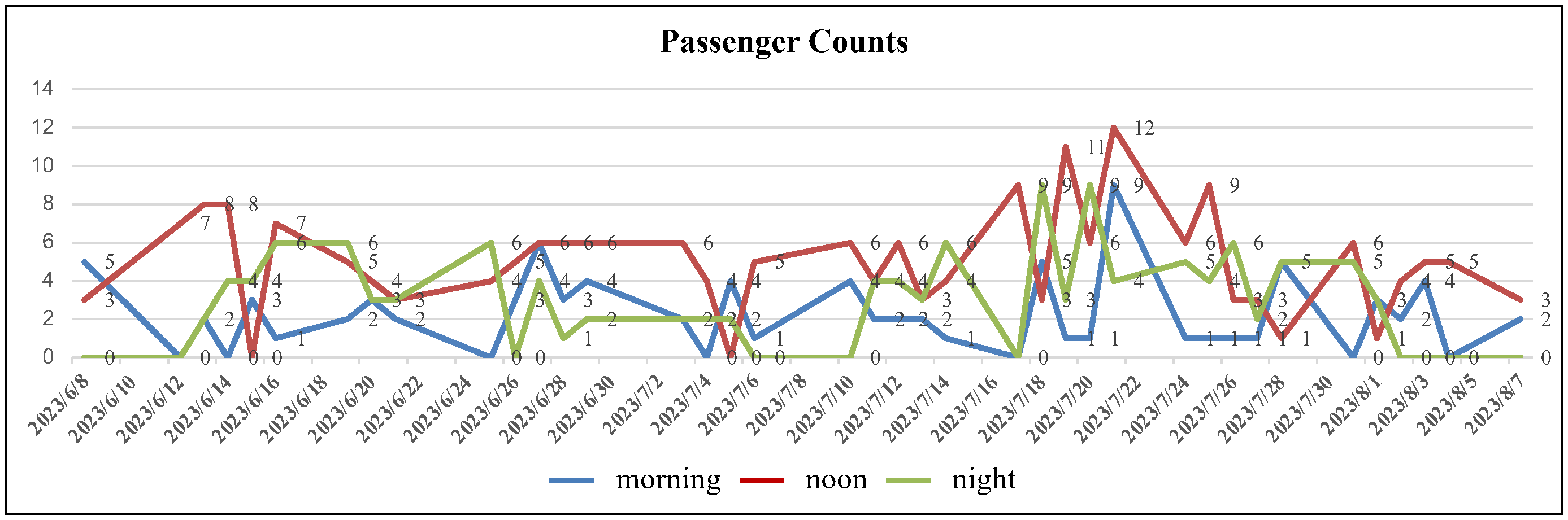}\label{subfig:quan_passenger}}
        \hspace{2.5pt} \subfigure[]{\includegraphics[width=0.33\linewidth]{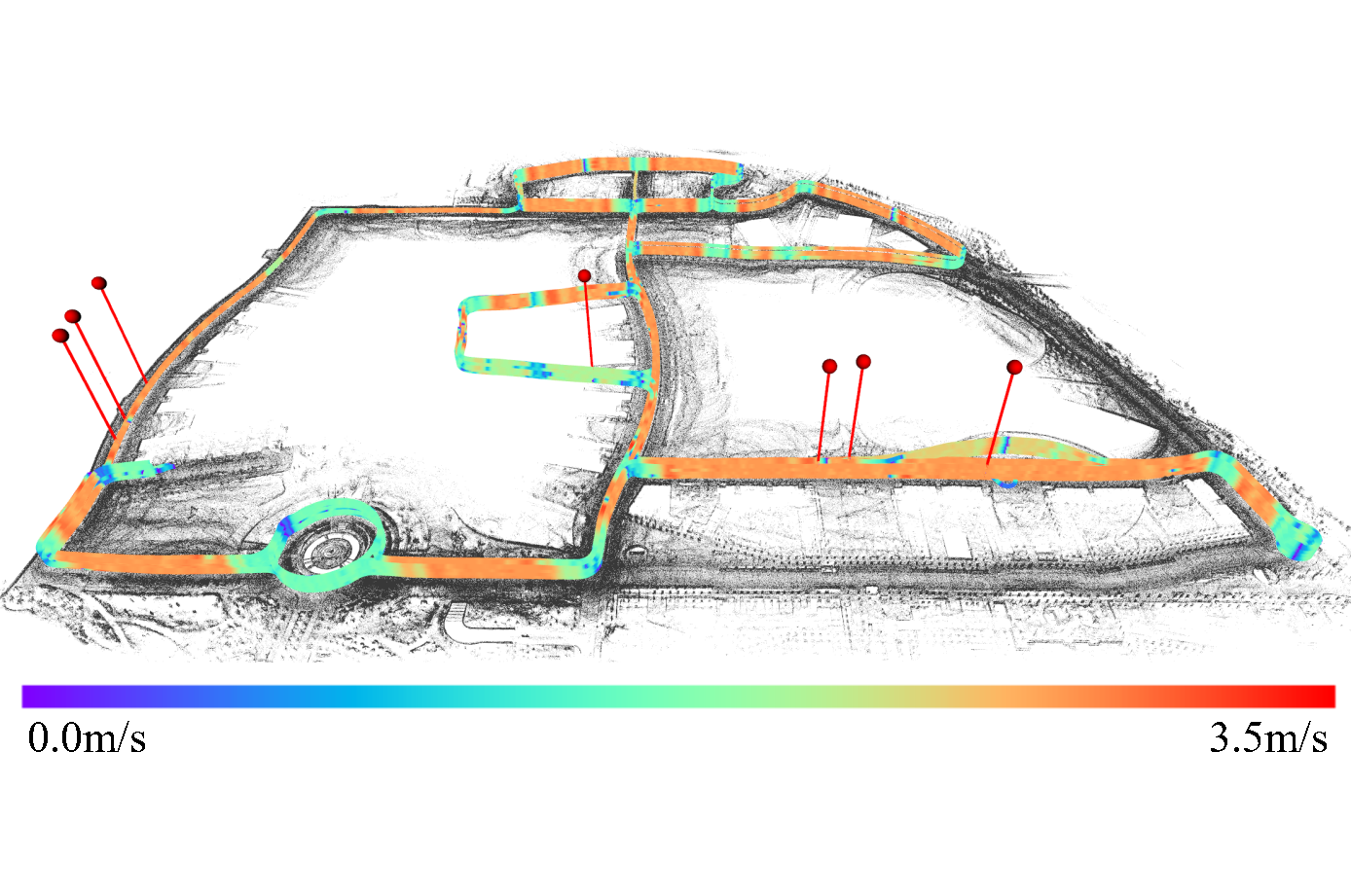}\label{subfig:quan_performances}}
    \end{minipage} 
    \caption{This figure demonstrates statistical results for the autonomous shuttle during its operation period. (a) It records passenger volumes at various times, with the periods distinguished based on the schedule in Table \ref{table:shuttle_schedule}. In total, the operation spanned 39 days, averaging 10.82 passengers per day.	The morning period resulted in 106 passengers, averaging 2.72 passengers.	The noon period recorded 197 passengers, averaging 5.05 passengers.	The night period produced 119 passengers, with an average of 3.05 passengers.	Figure (b) displays the speed distribution of the AV over the last 1000 kilometers traveled within the campus.
    }
    \label{fig:quan_results}
\end{figure*}

This section discusses the practical applications of our autonomous shuttle, ``Snow Lion'', on the HKUST (GZ) campus in Guangzhou, China. Between June 8, 2023, and August 7, 2023, "Snow Lion" operated for three periods each day, with two hours in each period, to enhance campus mobility. It followed predefined routes, covering various fixed locations. The cumulative travel distance for the AV amounted to 1,147 kilometers. Further information regarding these transportation tasks is presented in Table \ref{table:shuttle_schedule}. A selection of illustrative photos captured during the tasks is presented in Fig. \ref{fig:real_demos}. It is noteworthy that, in case of setbacks during autonomous shuttle operations, such as inevitable accidents or system errors, a security guard is consistently present within the AV. In the event of any autonomous navigation failure, the responsibility for vehicle control is promptly transferred to the security guard to ensure passenger safety. We anticipate reduced human intervention in the shuttle tasks. Performance evaluation is based on the frequency of human interventions and other pertinent criteria, all of which are automatically recorded in the ROS bag format. Detailed results are presented in Figure \ref{fig:quan_results} and Table \ref{table:quan_result}, where the symbol `BT' denotes the metric for measuring braking times. This metric registers an instance when the acceleration falls below $-1 \text{m}/\text{s}^2$. Additionally, `takeovers' signifies the average distance an autonomous shuttle covers before necessitating manual intervention by security personnel.

\begin{table}[t]
\renewcommand\arraystretch{1.15}   
\renewcommand\tabcolsep{8pt}  
\centering
\caption{Performances of autonomous shuttle tasks over the last 1000 kilometers traveled within the campus.}
\begin{tabular}{p{15.0mm}|l|p{15.0mm}|l}
    \hline
    Metric & Values & Metric & Values  \\ \hline
    Max. speed & 11.76 km/h & Avg. speed & 7.41 km/h \\ \hline
    Pct. speed $\geq 9$ km/h & 47.4 \% & Min. acc. & -1.83 $\text{m}/\text{s}^2$ \\ \hline
    Max. acc. & 1.12 $\text{m}/\text{s}^2$ & Min. jerk & -1.93 $\text{m}/\text{s}^3$ \\ \hline
    Max. jerk & 1.84 $\text{m}/\text{s}^3$ & BT (acc. $\leq -1 m/s^2$ ) & 0.77 km/time \\ \hline
    Takeovers & 13.34 km/time & - & - \\ \hline
\end{tabular}
\label{table:quan_result}
\vspace{-1.0em}
\end{table}

Analysis of the data in Table \ref{table:quan_result} reveals that the autonomous shuttle consistently maintains a comfortable driving performance, with an absolute deceleration of less than 1.83 $\text{m}/\text{s}^2$ and an absolute jerk value of less than 1.93 $\text{m}/\text{s}^2$ over the course of the last 1000 km of operations. Furthermore, upon reviewing the illustration in Figure \ref{subfig:quan_performances}, the autonomous shuttle consistently maintains a relatively high-speed performance, with speeds exceeding 9 km/h in in over 47.4\% of operational instances, but experiences speed reductions at intersections and during turning scenarios.

%% file: secs/lessons_conclusion.tex
\section{Lessons Learned and Conclusions}

Navigating an autonomous vehicle through a university campus, marked by a highly unpredictable environment with diverse traffic participants and an absence of traffic regulations, presents formidable challenges.	This section mainly discuss these challenges from a planning perspective, which predominantly emanate from several aspects.	

Firstly, from the standpoint of point-cloud mapping, localization instability may be unavoidable, particularly in scenarios with sparse point cloud features for mapping or when map features evolve over time (e.g., tree growth and leaf expansion affecting the effectiveness of an offline map). This failure may manifest, especially when the autonomous vehicle (AV) encounters a large truck, which can obstruct valuable environmental features for the lidar sensors. We address this issue by incorporating localization error information into the behavioral planning module.	When a significant localization error arises, it initiates a forced deceleration of the AV to alleviate the computational burden associated with the mapping process. This strategy is straightforward yet highly effective. Moreover, when the speed decreases, the mapping error rapidly recovers, restoring the normal autonomous driving status.	

\begin{figure}[t]
\centering \includegraphics[width=1.0\linewidth]{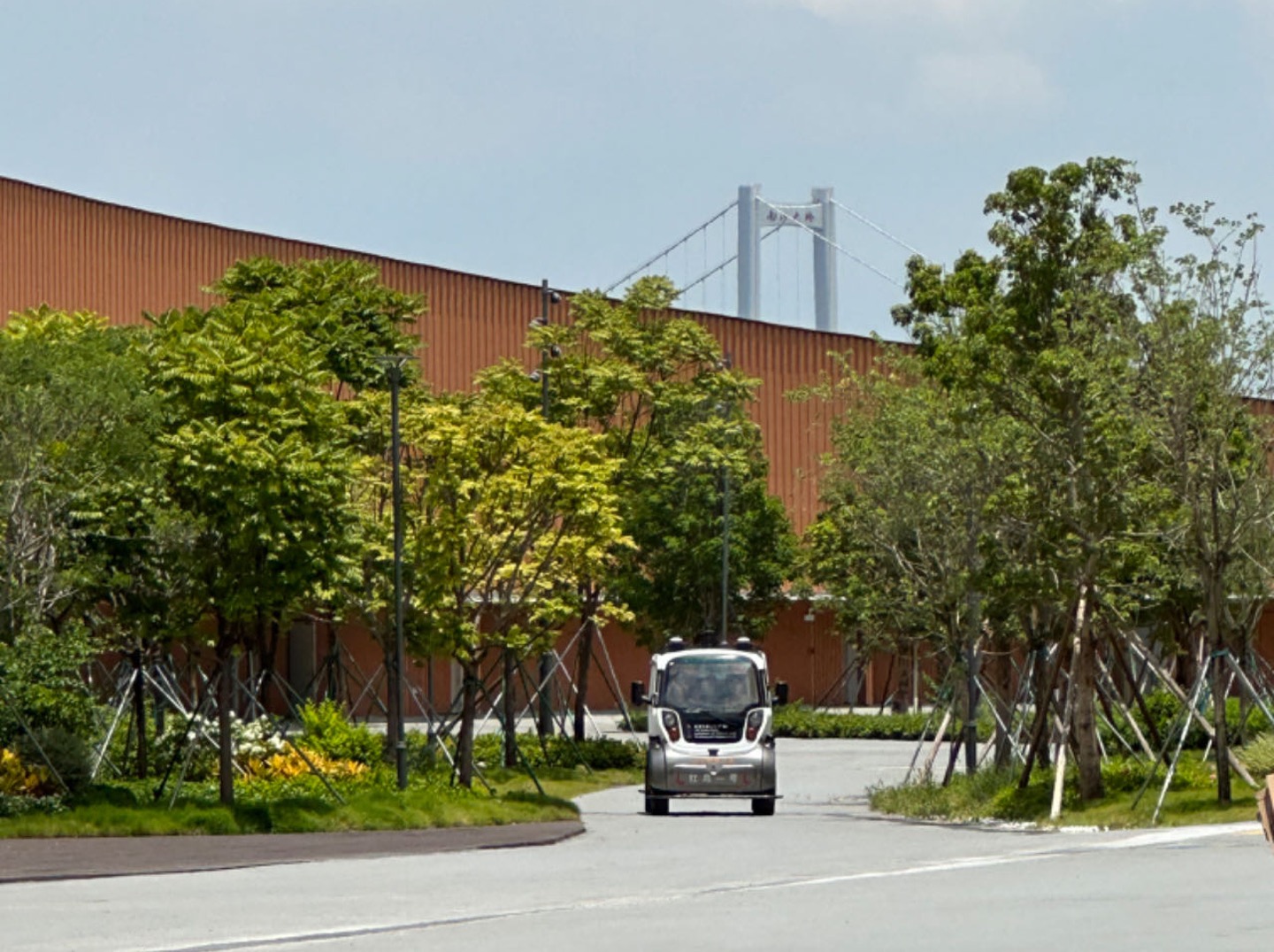}

\caption{This demonstrates situations where the existence of tree shade potentially affects the navigation performance of AVs, as it introduces noise into the perception algorithm, particularly when the trees sway in the wind, leading to fluctuations in the drivable area's boundary.
}
\label{fig:challenge_tree_shades}
\vspace{-0.0em}
\end{figure}

For results from object detection and trajectory prediction, it is doomed to be full of noise as the environment is full of uncertainty and less traffic regulations. We found that constant velocity model and a lane-keeping speed planning scheme is enough for a stable navigation performance. While it really matters to for the behavior decision process, for example, the lane-change manoeuvre, we solve this by adding certain delays in the state machine in the behavioral planning, making the navigation performance more stable. Another challenge arises from the tree shades along the roadway, depicted in Fig. \ref{fig:challenge_tree_shades}. These dynamic tree shades continually shift over time, often extending towards the center of the road. This situation complicates the ability to maintain a secure and consistent path for the autonomous vehicle, particularly in narrow paths. It also causes disruptions in the perception module, leading to distorted shape detection and potentially impacting speed recognition. Exploring more effective technologies to eliminate tree-related obstacles is crucial in resolving this issue.

Regarding the planning aspect, our investigation revealed that clear planning behavior for other road users and the understanding of their intentions are crucial factors for user-friendly autonomous driving. This includes recognizing whether other agents are aware of their interaction with the AV or understand that the AV acknowledges their intentions. Ensuring clarity in these interactions is vital to prevent sudden braking and ensure safety for the AV.
Despite the slowdown strategy (\ref{eq:vlimit_pedestrian}) implemented in our planning process effectively mitigating these concerns, resulting in an average of 13.34 kilometers for each manual takeover during the operational period, nevertheless, these challenges persist and warrant further investigation in future studies.

Throughout the operational phase, two additional concerns affecting autonomous navigation performance are evident. Firstly, the intention detection module faces challenges in effectively operating within a non-fixed road network environment. In such an environment, traffic agents often exhibit unexpected maneuvers, such as overtaking, jaywalking, and more, owing to the absence of fully-compliant lane constraints in these scenarios. Current prediction methods heavily depend on this data, and consequently, frequently encounter difficulties in accurately forecasting the movements of other agents. In our research, we employ manually defined observation areas to tackle these challenges, especially in intricate intersections and unstructured regions. However, this strategy still places a substantial workload on fine-tuning navigation performance for new scenarios. Another concern pertains to responsibility. In the event of an accident involving an AV, liability requires the identification of a responsible party. This poses challenges for the proliferation of autonomous driving technologies and underscores the need for sound legal regulations.

Overall, this paper presents the design of an autonomous shuttle system, highlighting its quantitative performance and discussing challenges in scenarios with limited traffic regulation. The proposed system improves campus mobility, facilitating the transport of 235 passengers over 39 days of operation. In future research, we will endeavor to reduce takeover times and address the previously mentioned challenges.


%% file: main.bbl
\begin{thebibliography}{1}
\providecommand{\url}[1]{#1}
\csname url@samestyle\endcsname
\providecommand{\newblock}{\relax}
\providecommand{\bibinfo}[2]{#2}
\providecommand{\BIBentrySTDinterwordspacing}{\spaceskip=0pt\relax}
\providecommand{\BIBentryALTinterwordstretchfactor}{4}
\providecommand{\BIBentryALTinterwordspacing}{\spaceskip=\fontdimen2\font plus
\BIBentryALTinterwordstretchfactor\fontdimen3\font minus \fontdimen4\font\relax}
\providecommand{\BIBforeignlanguage}[2]{{%
\expandafter\ifx\csname l@#1\endcsname\relax
\typeout{** WARNING: IEEEtran.bst: No hyphenation pattern has been}%
\typeout{** loaded for the language `#1'. Using the pattern for}%
\typeout{** the default language instead.}%
\else
\language=\csname l@#1\endcsname
\fi
#2}}
\providecommand{\BIBdecl}{\relax}
\BIBdecl

\bibitem{liu2021role}
T.~Liu, Q.~hai Liao, L.~Gan, F.~Ma, J.~Cheng, X.~Xie, Z.~Wang, Y.~Chen, Y.~Zhu, S.~Zhang \emph{et~al.}, ``The role of the hercules autonomous vehicle during the covid-19 pandemic: An autonomous logistic vehicle for contactless goods transportation,'' \emph{IEEE Robotics \& Automation Magazine}, vol.~28, no.~1, pp. 48--58, 2021.

\bibitem{zhou2018voxelnet}
Y.~Zhou and O.~Tuzel, ``Voxelnet: End-to-end learning for point cloud based 3d object detection,'' in \emph{2018 IEEE/CVF Conference on Computer Vision and Pattern Recognition}, 2018, pp. 4490--4499.

\bibitem{jiao2019novel}
J.~Jiao, Q.~Liao, Y.~Zhu, T.~Liu, Y.~Yu, R.~Fan, L.~Wang, and M.~Liu, ``A novel dual-lidar calibration algorithm using planar surfaces,'' in \emph{2019 IEEE Intelligent Vehicles Symposium (IV)}.\hskip 1em plus 0.5em minus 0.4em\relax IEEE, 2019, pp. 1499--1504.

\bibitem{shan2018lego}
T.~Shan and B.~Englot, ``Lego-loam: Lightweight and ground-optimized lidar odometry and mapping on variable terrain,'' in \emph{2018 IEEE/RSJ International Conference on Intelligent Robots and Systems (IROS)}.\hskip 1em plus 0.5em minus 0.4em\relax IEEE, 2018, pp. 4758--4765.

\bibitem{valls2018design}
M.~I. Valls, H.~F. Hendrikx, V.~J. Reijgwart, F.~V. Meier, I.~Sa, R.~Dub{\'e}, A.~Gawel, M.~B{\"u}rki, and R.~Siegwart, ``Design of an autonomous racecar: Perception, state estimation and system integration,'' in \emph{2018 IEEE international conference on robotics and automation (ICRA)}.\hskip 1em plus 0.5em minus 0.4em\relax IEEE, 2018, pp. 2048--2055.

\bibitem{chen2022efficient}
Y.~Chen, R.~Xin, J.~Cheng, Q.~Zhang, X.~Mei, M.~Liu, and L.~Wang, ``Efficient speed planning for autonomous driving in dynamic environment with interaction point model,'' \emph{IEEE Robotics and Automation Letters}, vol.~7, no.~4, pp. 11\,839--11\,846, 2022.

\bibitem{cheng2022real}
J.~Cheng, Y.~Chen, Q.~Zhang, L.~Gan, C.~Liu, and M.~Liu, ``Real-time trajectory planning for autonomous driving with gaussian process and incremental refinement,'' in \emph{2022 International Conference on Robotics and Automation (ICRA)}.\hskip 1em plus 0.5em minus 0.4em\relax IEEE, 2022, pp. 8999--9005.

\bibitem{rajamani2011vehicle}
R.~Rajamani, \emph{Vehicle dynamics and control}.\hskip 1em plus 0.5em minus 0.4em\relax Springer Science \& Business Media, 2011.

\end{thebibliography}
